\documentclass[conference]{IEEEtran}
\IEEEoverridecommandlockouts
\usepackage{cite}
\usepackage{amsmath,amssymb,amsfonts}
\usepackage{algorithmic}
\usepackage{graphicx}
\usepackage{textcomp}
\usepackage{longtable}
\usepackage{mathtools}
\usepackage{multirow}
\usepackage[utf8]{inputenc}
\usepackage{amsmath}
\usepackage{booktabs}
\usepackage{mathabx} 

\def\BibTeX{{\rm B\kern-.05em{\sc i\kern-.025em b}\kern-.08em
    T\kern-.1667em\lower.7ex\hbox{E}\kern-.125emX}}

\newcommand{\norm}[1]{\left\lVert#1\right\rVert}
\graphicspath{ {images/} }

\begin{document}

\title{An Overwiew of Datatype Quantization Techniques for Convolutional Neural Networks
}

\author{\IEEEauthorblockN{Ali Athar}
\IEEEauthorblockA{\textit{Department of Electrical and Computer Engineering} \\
\textit{Technical University of Munich}\\
Munich, Germany \\
ali.athar@tum.de}
}

\maketitle

\begin{abstract}
Convolutional Neural Networks (CNNs) are becoming increasingly popular due to their superior performance in the domain of computer vision, in applications such as objection detection and recognition. However, they demand complex, power-consuming hardware which makes them unsuitable for implementation on low-power mobile and embedded devices. In this paper, a description and comparison of various techniques is presented which aim to mitigate this problem. This is primarily achieved by quantizing the floating-point weights and activations to reduce the hardware requirements, and adapting the training and inference algorithms to maintain the network's performance.  
\end{abstract}


\section{Introduction}
In recent years, CNNs have proven to be much more effective in computer vision tasks as compared to existing conventional approaches. One drawback of CNNs, however, is their complexity in terms of the number of parameters and computations. For this reason, neural networks are typically applied on high-powered Graphical Processing Units (GPUs) with huge memory space. As an example, AlexNet\cite{NIPS2012_4824} contains approximately 61M parameters, and performs 1.5B high-precision operations in order to classify a single image. This makes it prohibitively expensive to train CNNs on resource-constrained mobile or embedded devices. 

Significant research has therefore been undertaken to discover ways to reduce the memory and computational resources required by CNNs. The principal approach has been to quantize the high precision network parameters by reducing their word lengths, so as to reduce (a) the memory bandwidth and storage, (b) memory access and floating-point computations, and (c) power consumption. However, arbitrary quantization of all network parameters significantly degrades the network's performance. To avoid this, the back-propagation algorithm has to be modified, and the degree of quantization has to be carefully optimized to achieve a balance between network compression and performance.

In this paper, an analysis of some works that aim to compress CNNs is presented. The organization is as follows: section II provides a brief round-up of many network compression techniques. In section III, the most recent and promising of these approaches are discussed in detail. Section IV compares the results produced by these approaches and how they compare to existing state-of-the-art networks in terms of inference performance and memory requirements.

\section{Related Work}

The various techniques that have been proposed to reduce the memory and computational footprint of CNNs can be roughly categorized in one of two groups:

\subsection{Compressing Pretrained Networks} One of the earliest works in this area was Optimal Brain Surgeon\cite{hassibi1993second}, which pruned the network (i.e. remove unnecessary weights) by using a metric based on the Hessian of the loss function. More recently, Han \textit{et al.}\cite{han2015deep} introduced Deep Compression, which involved a three-stage pipeline that (1) pruned, (2) quantized and partially retrained, and (3) compressed the resulting weights using Huffman Coding. Denton \textit{et al.}\cite{denton2014exploiting} noted that there are redundancies within the parameters of each layer, and proposed using matrix factorization methods to reduce the number of parameters.

Chen \textit{et al.}\cite{chen2015compressing} proposed HashedNets, wherein a low-cost hashing function is used to randomly group weights into hash buckets, and all weights in a single bucket share a single parameter value. Vanhoucke \textit{et al.}\cite{vanhoucke2011improving} quantized the weights and activations of a pretrained network to a fixed-point length of 8-bits. As a refinement of this approach, Anwar \textit{et al.}\cite{anwar2015fixed} apply layer-specific quantization based on the sensitivity of each layer and an $L_2$-error minimization technique. Lin \textit{et al.}\cite{lin2016fixed} adopt a similar approach, but use an optimization strategy based on minimizing the signal-to-quantization-noise-ratio (SQNR) to determine the bit-width for each layer. Shin \textit{et al.}\cite{shin2017fixed} retrain deep networks using a variable quantization step-size based on the $L_2$-distance between the fixed-point and floating-point weights. Hwang \textit{et al.}\cite{hwang2014fixed} on the other hand, enforce ternary weights while quantizing (-1, 0 and +1), and then retrain the network after quantization to regain inference accuracy.

\subsection{Training Quantized Networks} A comparatively smaller set of works focuses on training quantized networks from scratch. Lin \textit{et al.}\cite{li2016ternary} introduce Ternary Weights Networks (TWNs), wherein all weights have values -1, 0 or +1. By assuming that all weights are uniformly or normally distributed, they were able to determine the thresholds for quantization. Meng \textit{et al.}\cite{meng2017two} build upon this, but utilize the two-bit weights fully by allowing four possible weight values (-2, -1, +1, +2), and learning the optimal scaling factor for each kernel.

Courbariaux \textit{et al.}\cite{courbariaux2015binaryconnect} introducted BinaryConnect, wherein they restrict all weight values to either -1 or +1, and further argue that binarization, in addition to compressing the network, also provides a form of regularization. Rastegari \textit{et al.}\cite{rastegari2016xnor} came up with XNOR-nets, in which they not only quantize the kernel weights, but also the network inputs. Furthermore, they formulate the convolution as a series of XNOR operations to speed up the computations.

\section{Datatype Quantization}
In this section, some of the contemporary works mentioned in the previous section are described in more detail. The works are chosen so as to provide a holistic overview of all the approaches that yield appreciable results and offer interesting theoretical perspectives.

\subsection{Compressing Pretrained Networks}
Generally, compressing pretrained networks involves one or both of the following steps: (1) Determining the quantization parameters, namely the step-size and bitwidth. These may be preset based on an assumption, or they may be computed analytically; (2) Retraining the network after quantization in order to partially recover the loss in inference accuracy.

\vspace{5pt}
\subsubsection{Variable bit-width (Lin et al.)}
In \cite{lin2016fixed}, the quantized network is not retrained, but both quantization parameters are determined analytically. Assuming that the weights in a CNN are normally distributed, the quantization parameters are determined by minimizing the SQNR. To this end, an approximate linear expression for SQNR is derived. The relationship between the floating-point and quantized products of the weight and activation can be given as:

\vspace{-6pt}
\begin{align*}
\widetilde{w} \cdot \widetilde{a} & =(w+n_{w}) \cdot (a+n_{a}) \simeq w\cdot a + w\cdot n_{a} + n_{w}\cdot a
\end{align*}

where $w$ and $a$ are the floating-point weights and activations (input from previous layer) respectively, $\widetilde{w}$ and $\widetilde{a}$ are the quantized weights and activations, and $n_{w}$ and $n_{a}$ are the quantization errors associated with the weight and activation respectively. The above approximation holds if $|n_{a}| \ll |a|$ and $|n_{w}| \ll |w|$. Here, it is apparent that $(w\cdot n_{a} + n_{w}\cdot a)$ is the overall quantization noise when taking the product of $w$ and $a$. From this, it is deduced that the SQNR of $w\cdot a$, $\gamma_{w\cdot a}$ is related to the individual SQNRs of $w$ and $a$ ($\gamma_{w}$ and $\gamma_a$) as follows:

\vspace{-8pt}
\begin{align*}
\frac{1}{\gamma_{w\cdot a}} &= \frac{\mathbf{E}(w\cdot n_{a} + n_{w}\cdot a)^2}{\mathbf{E}(w\cdot a)^2} \simeq \frac{1}{\gamma_{w}} + \frac{1}{\gamma_{a}}
\end{align*} 

\vspace{3pt}
This is an important linear relation that can be extended to multiple layers. The final output SQNR $\gamma_{output}$ for a CNN with $n$ layers is therefore:

\vspace{-4pt}
\begin{equation}\label{eq:1}
\gamma_{output} = \frac{1}{\gamma_{w^{(0)}}} + \frac{1}{\gamma_{a^{(0)}}} + ... + \frac{1}{\gamma_{w^{(n)}}} + \frac{1}{\gamma_{a^{(n)}}}
\end{equation}
\vspace{3pt}

The overall SQNR is thus simply the harmonic mean of the individual SQNRs at each step. Separately, is shown that in general, $10\log{\gamma}=\kappa\cdot \beta$, where $\kappa$ is the quantization efficiency which is dependent on the type of distribution assumed, and $\beta$ is the bitwidth. Combining this relation with \eqref{eq:1}, it is trivial to formulate an optimization problem that minimizes the bitwidth $\beta_i$ for each layer $i$, with the constraint that the SQNR $\gamma_i$ does not fall below a certain threshold. 

\vspace{5pt}
\subsubsection{Ternary Weights (Hwang et al.)}
In \cite{hwang2014fixed}, ternary weights $+1, 0, -1$ are enforced, so the bitwidth is 2. Given a pretrained network with floating-point weights, the quantization is carried out in three steps as follows:

\begin{enumerate}
\item Compute an initial step-size $\Delta_{init}$ for the entire network using an $L_2$-error minimization approach similar to Lloyd-Max quantization.
\item Quantize the weights in all layers of the network using step-size $\Delta_{init}$ i.e. $\Delta_i = \Delta_{init}, \forall i \in [1,n]$, where $n$ is the number of layers.
\item Fine-tune the step-size layer-wise. Starting from the quantizer in the first layer i.e. $i = 1$:
\begin{enumerate} 
\item Try several step-sizes $\Delta_i'$ close to the initial step-size, and for each $\Delta_i'$, calculate the $L_2$-error at the output. Set $\Delta_i = \Delta_i^*$, where $\Delta_i^*$ is the $\Delta_i'$ that achieved the lowest $L_2$-error.
\item Increment $i$ by 1 and repeat steps (a) and (b) until $i=n$.
\end{enumerate}
\end{enumerate} 

The quantized network is then retrained to fine-tune the weights. To this end, the standard back-propagation algorithm with Stochastic Gradient Descent (SGD) and mini-batches is used, but with two modifications:

\begin{enumerate}
\item The forward and back-propagation is carried out using the quantized weights, but the parameter update is still applied to the floating-point weights. This is because the amount of update is usually very small at a single step. The updated floating-point weights are then quantized for further use.
\item The floating-point weights are used for calculating the derivative of the activation function during back-propagation. This is because the derivative of the sigmoid activation function often vanishes to 0 when the function value only has three quantization levels.
\end{enumerate}

\subsection{Training Quantized Networks}
These works implement a quantized network by training it from scratch.

\vspace{3pt}
\subsubsection{XNOR-Net (Rastegari et al.)}
In \cite{rastegari2016xnor}, XNOR-Nets are introduced which binarize both  the weights and the inputs to the network. For a given 3D convolution filter $W$ with $n$ floating-point weights, let the binarized filter be $\widetilde{W} =\alpha B$ (where $\alpha$ is a scaling factor and $B \in \{+1,-1\}^n$). Attempting to minimize the $L_2$-error between the floating-point and quantized weights yields the familiar optimization problem:

\vspace{-8pt}
\begin{align*}
\alpha^*, B^* &= 
\underset{\alpha > 0, B \in \{+1,-1\}^n}{\mathrm{argmin}} \norm{W-\alpha B}_2^2 \\
&= \underset{\alpha > 0, B \in \{+1,-1\}^n}{\mathrm{argmin}} (W^TW - 2\alpha W^TB + \alpha^2 B^TB)
\end{align*}

Since $W$ is a known variable and $W^TW$ and $B^TB$ are constants, the problem simplifies to $B^* = \underset{B}{\mathrm{argmax}} (W^TB)$. The solution to this is simply $B^* = sign(W)$ i.e. $B_i$ is +1 when $W_i$ is positive, and vice versa. Solving for $\alpha^*$ in the above equation and substituting the expression for $B^*$, the following solution is obtained:

\vspace{-6pt}
\begin{equation}
\alpha^* = \frac{1}{n} \norm{W}_1
\end{equation}

This shows that the optimal step-size $\alpha^*$ is simply the mean of the absolute values of all the individual weights in $W$.
Insofar as binarization of inputs is concerned, it should be recalled that a convolution is simply a series of dot-product and shift operations. It follows that if both weight and input matrices to a layer are purely binary, the operation can be implemented using highly efficient XNOR-bitcounting operations\cite{courbariaux2016binarized}. In order to binarize the input $I$ that is to be convolved with a weight kernel $W$, the objective is to minimize the $L_2$-error for every single dot-product operation between $W$ and the sub-matrix $X$ (belonging to $I$) being operated on. We denote binarized $X$ as $\widetilde{X}$ and express it in a similar way to $\widetilde{W}$: $\widetilde{X} = \beta H$, where $\beta$ is a scaling factor and $H \in \{+1,-1\}^n$. The optimal value of $H$, $H^*$, and the optimal value of $B$, $B^*$ is found to be:

%

\vspace{-8pt}
\begin{equation}\label{eq:BB3}
H^* = sign(X),\hspace{8pt} \beta^* = \frac{1}{n} \norm{X}_1
\end{equation}

Interestingly, this result has the same form as that obtained for $\alpha^*$ and $B^*$. The fully binarized network is then trained using the standard SGD in conjunction with momentum-based or ADAM learning rate adaptation, but there are two noteworthy changes: (1) the floating-point weights are still used for parameter updating (for the same reason as Hwang et al.\cite{hwang2014fixed}), and (2) for back-propagation, the following derivative of the $sign()$ function is used\cite{courbariaux2016binarized}:

\vspace{-6pt}
\begin{equation}
\frac{\partial sign(r)}{\partial r} = 
\begin{cases}
r    & |r| \leq 1  \\
0    & otherwise
\end{cases}
\end{equation}

\vspace{5pt}
\subsubsection{Ternary Weights (Li et al.)}
In \cite{li2016ternary}, a ternary network is implemented i.e. the bitwidth is fixed to two. The step-size, $\Delta$ is determined by attempting to minimize the Euclidean distance between the floating-point weights $w_i$ and quantized weights $\widetilde{w}_i$. The following quantization function is used:

\vspace{-8pt}
\begin{equation}\label{eq:BT1}
\widetilde{w_i} = 
\begin{cases}
+\alpha,  & w_i > \Delta      \\
0,        & \left|w_i\right| \leq \Delta \\
-\alpha,  & w_i < -\Delta
\end{cases}
\end{equation}

where $i$ is the weight index within a particular filter, and $\alpha$ is simply a scaling factor that is applied to the ternary weights $+1, 0, -1$. The optimization problem can be formulated as:

\vspace{-5pt}
\begin{equation}\label{eq:BT2}
\alpha^* = \underset{\alpha}{\mathrm{arg min}} \norm{w - \widetilde{w}}_2^2
\end{equation}
\vspace{-5pt}

Substituting \eqref{eq:BT1} into \eqref{eq:BT2} and expanding the squared term, the problem can be rewritten as:

\begin{equation}\label{eq:BT3}
\alpha^*, \Delta^* = \underset{\alpha\geq 0,\Delta\geq 0}{\mathrm{arg min}}(|\bold{I}_{\Delta}|\alpha^2 - 2(\sum\limits_{i\in \bold{I}_{\Delta}} |w_i| )\alpha + c_{\Delta})
\end{equation}

\vspace{3pt}
where $\bold{I}_{|\Delta|} = \{ i\hspace{1pt} |\hspace{1pt} |w_{i}| \leq \Delta \} $, $|\bold{I}_{\Delta}|$ denotes the number of elements in $\bold{I}_\Delta$, and $c_\Delta = \sum\limits_{i \in \bold{I}_{\Delta}^c}w_i^2$. The solution for $\alpha^*$ is:

\vspace{4pt}
\begin{equation}\label{eq:BT4}
\alpha^* = \frac{1}{|\bold{I}_\Delta|} \sum\limits_{i\in \bold{I}_\Delta} |w_i|
\end{equation}

i.e. the mean of all floating-point weights in $\bold{I}_\Delta$. Substituting \eqref{eq:BT4} into \eqref{eq:BT3} gives:

\vspace{-8pt}
\begin{equation}\label{eq:BT5}
\Delta^* = \underset{\Delta > 0}{\mathrm{argmin}} \frac{1}{|\bold{I}_\Delta|} \sum\limits_{i\in \bold{I}_\Delta} w_i^2
\end{equation}  
\vspace{3pt}

In order to solve this problem easily, a simplifying assumption is made similar to \cite{lin2016fixed} that all $w_i$ are either: (1) uniformly distributed, in which case $\Delta^* = \frac{2}{3} \mathbf{E}{|w|}$, or (2) normally distributed with zero mean, in which case $\Delta^* = 0.75\mathbf{E}{|w|}$.

To train the network, SGD is used in conjunction with Batch Normalization and momentum-based learning rate scaling. The quantized ternary weights are used during both forward and back-propagation, but for parameter update, the high-precision floating-point weights are used.

\section{Experimental Results}

The results for the networks from subsection III-B are given in Table \ref{tab:table1}. The top-1 and top-5 accuracy percentages on the ImageNet database\cite{ImageNet} for three quantization schemes are given. The model compression (i.e. the factor by which the memory footprint of the network is reduced in comparison to the non-quantized network) can be explained by the fact that floating-point values are normally represented as 32-bit variables. In ternary and binary-weighted networks however, the weights are represented using 2-bit and 1-bit variables respectively, thereby resulting in 16x and 32x model compression. 

\begin{table}[h!]
  \begin{center}
  \caption{Validation accuracies (\%) on ImageNet}
    \begin{tabular}{l|c|c|c|c|c} 
    \hline
      \multirow{2}{*}{\textbf{Network Type}} & \multicolumn{2}{c|}{\textbf{ResNet-18}\cite{he2016deep}} & \multicolumn{2}{c|}{\textbf{AlexNet}\cite{NIPS2012_4824}} & \textbf{Model} \\\cline{2-5}
      & top-1 & top-5 & top-1 & top-5 & \textbf{Compression}\\[1pt]
      \hline
      Ternary Weight\cite{li2016ternary} & 61.8       & 84.2 & -    & -    &  16\\
      Binary Weight\cite{rastegari2016xnor}$^1$  & 60.8       & 83.0 & 56.8 & 79.4 &  32\\
      XNOR-Net\cite{rastegari2016xnor}$^2$        & 51.2       & 73.2 & 44.2 & 69.2 &  32\\
      \hline
      Full Precision  & 69.3       & 89.2 & 56.6 & 80.2 &  1\\
    \end{tabular}
    \label{tab:table1}
  \end{center}
\end{table}
\vspace{-8pt}
\footnotesize{$^1$ Binarized weight kernels, $^2$ Binarized inputs and weight kernels}
\vspace{4pt}
\normalsize

\begin{figure}[h]
\centering
\includegraphics[width=0.8\linewidth]{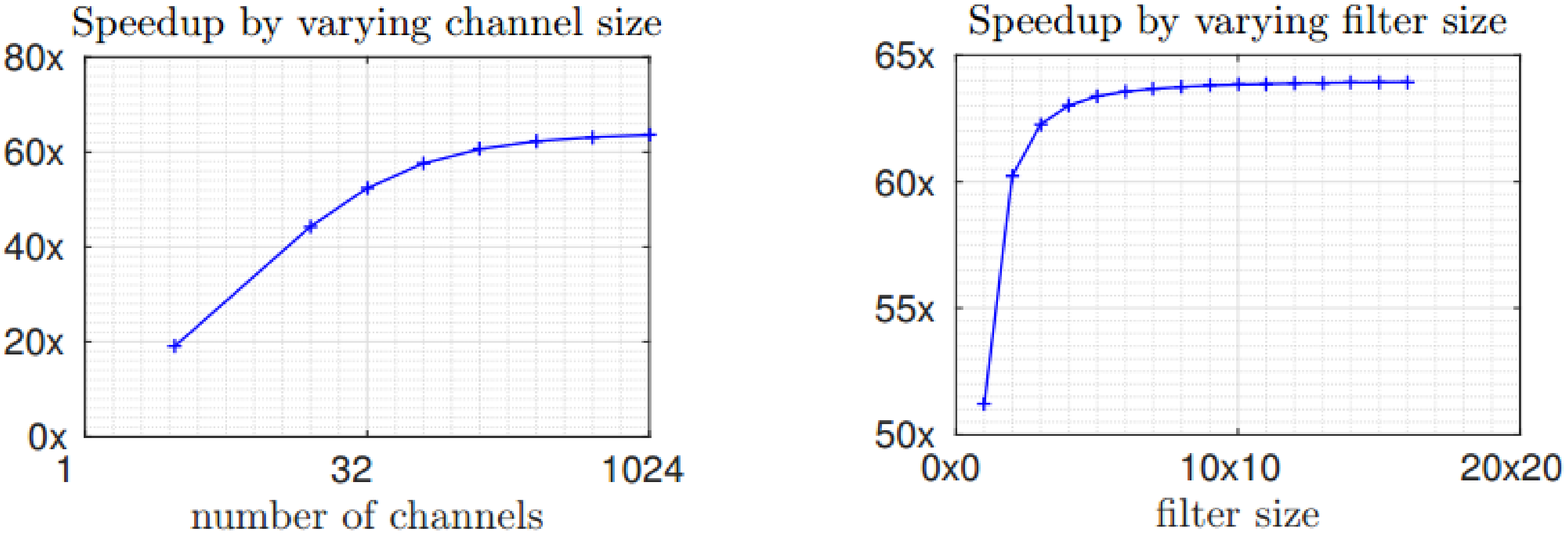}
\caption{Speedup on XNOR-net\cite{rastegari2016xnor} for various filter sizes and number of channels}
\label{fig:speedup}
\end{figure}

With regard to speedup in CPU processing time, Rastegari \textit{et al.} \cite{rastegari2016xnor} offer the most promising results. Consider that in a standard convolution, the total number of operations is $cN_WN_I$, where $c$ is the number of channels, and $N_W$ and $N_I$ are the number of elements in a single channel of the weight kernel and input matrix respectively. Using their proposed binary approximation however, convolution can be performed with $cN_WN_I$ binary operations and $N_I$ non-binary operations. Since modern CPUs can perform 64 binary operations per clock cycle, the speedup $S = \frac{cN_WN_I}{\frac{1}{64}cN_WN_I+N_I} = \frac{64cN_W}{cN_W+64}$. The actual speedup achieved for various filter sizes and number of channels is given in Figure \ref{fig:speedup}.

\vspace{-4pt}

Figure 1 summarizes the results from the work of \textit{Lin et al.}\cite{lin2016fixed} which quantizes pretrained networks. It shows the top-5 error rate on ImageNet using AlexNet architecture; the improvement gain by using their layer-wise optimization approach as opposed to uniformly quantizing all the layers can also be seen. As comparison, the top-5 error rate of the full-precision, floating-point AlexNet\cite{NIPS2012_4824} is 17\%

\begin{figure}[h]
\centering
\includegraphics[width=0.5\linewidth]{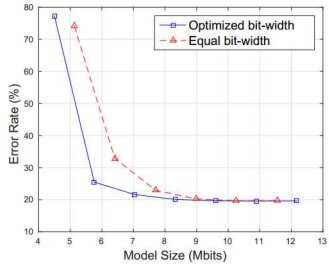}
  \caption{Top-5 error rates on ImageNet using quantization scheme proposed by \textit{Lin et al.}\cite{lin2016fixed}}
  \label{fig:fixedpoint}
\end{figure}

Lastly, the ternary quantization scheme for converting pretrained networks proposed by \textit{Hwang et al.} resulted in a classification error rate of 1.08\% on the MNIST database of handwritten digits\footnote{MNIST: http://yann.lecun.com/exdb/mnist/}. This error rate is comparable to that observed by full-precision networks implemented with the LeNet-4 architecture\cite{lecun1998gradient}.

\section*{Conclusion}

Quantization of CNNs enables them to be deployed on low-power, resource-constrained devices. In this paper, an overview of several network quantization approaches was given, with a more detailed description of some. Even though floating-point weights are still used during the weight update phase of back-propagation, the need for floating-point multiplications is eliminated. This is beneficial if CNNs are to be implemented on microcontrollers, since the latter often have limited floating-point computation capability, and similarly beneficial for FPGA-based implementations since floating-point multipliers are costly in terms of the number of slices required. In the best case, if all weights and inputs are binary, multiply-accumulate operations are replaced by 1-bit XNOR-count operations which are extremely cheap to implement on hardware. 

\bibliography{datatype_quantization_in_cnns}{}

\begin{thebibliography}{10}
\providecommand{\url}[1]{#1}
\csname url@samestyle\endcsname
\providecommand{\newblock}{\relax}
\providecommand{\bibinfo}[2]{#2}
\providecommand{\BIBentrySTDinterwordspacing}{\spaceskip=0pt\relax}
\providecommand{\BIBentryALTinterwordstretchfactor}{4}
\providecommand{\BIBentryALTinterwordspacing}{\spaceskip=\fontdimen2\font plus
\BIBentryALTinterwordstretchfactor\fontdimen3\font minus
  \fontdimen4\font\relax}
\providecommand{\BIBforeignlanguage}[2]{{%
\expandafter\ifx\csname l@#1\endcsname\relax
\typeout{** WARNING: IEEEtran.bst: No hyphenation pattern has been}%
\typeout{** loaded for the language `#1'. Using the pattern for}%
\typeout{** the default language instead.}%
\else
\language=\csname l@#1\endcsname
\fi
#2}}
\providecommand{\BIBdecl}{\relax}
\BIBdecl

\bibitem{NIPS2012_4824}
A.~Krizhevsky, I.~Sutskever, and G.~E. Hinton, ``Imagenet classification with
  deep convolutional neural networks,'' in \emph{Advances in Neural Information
  Processing Systems 25}, F.~Pereira, C.~J.~C. Burges, L.~Bottou, and K.~Q.
  Weinberger, Eds.\hskip 1em plus 0.5em minus 0.4em\relax Curran Associates,
  Inc., 2012, pp. 1097--1105.

\bibitem{hassibi1993second}
B.~Hassibi and D.~G. Stork, ``Second order derivatives for network pruning:
  Optimal brain surgeon,'' in \emph{Advances in neural information processing
  systems}, 1993, pp. 164--171.

\bibitem{han2015deep}
S.~Han, H.~Mao, and W.~J. Dally, ``Deep compression: Compressing deep neural
  networks with pruning, trained quantization and huffman coding,'' \emph{arXiv
  preprint arXiv:1510.00149}, 2015.

\bibitem{denton2014exploiting}
E.~L. Denton, W.~Zaremba, J.~Bruna, Y.~LeCun, and R.~Fergus, ``Exploiting
  linear structure within convolutional networks for efficient evaluation,'' in
  \emph{Advances in Neural Information Processing Systems}, 2014, pp.
  1269--1277.

\bibitem{chen2015compressing}
W.~Chen, J.~Wilson, S.~Tyree, K.~Weinberger, and Y.~Chen, ``Compressing neural
  networks with the hashing trick,'' in \emph{International Conference on
  Machine Learning}, 2015, pp. 2285--2294.

\bibitem{vanhoucke2011improving}
V.~Vanhoucke, A.~Senior, and M.~Z. Mao, ``Improving the speed of neural
  networks on cpus,'' in \emph{Proc. Deep Learning and Unsupervised Feature
  Learning NIPS Workshop}, vol.~1, 2011, p.~4.

\bibitem{anwar2015fixed}
S.~Anwar, K.~Hwang, and W.~Sung, ``Fixed point optimization of deep
  convolutional neural networks for object recognition,'' in \emph{Acoustics,
  Speech and Signal Processing (ICASSP), 2015 IEEE International Conference
  on}.\hskip 1em plus 0.5em minus 0.4em\relax IEEE, 2015, pp. 1131--1135.

\bibitem{lin2016fixed}
D.~Lin, S.~Talathi, and S.~Annapureddy, ``Fixed point quantization of deep
  convolutional networks,'' in \emph{International Conference on Machine
  Learning}, 2016, pp. 2849--2858.

\bibitem{shin2017fixed}
S.~Shin, Y.~Boo, and W.~Sung, ``Fixed-point optimization of deep neural
  networks with adaptive step size retraining,'' \emph{arXiv preprint
  arXiv:1702.08171}, 2017.

\bibitem{hwang2014fixed}
K.~Hwang and W.~Sung, ``Fixed-point feedforward deep neural network design
  using weights+ 1, 0, and- 1,'' in \emph{Signal Processing Systems (SiPS),
  2014 IEEE Workshop on}.\hskip 1em plus 0.5em minus 0.4em\relax IEEE, 2014,
  pp. 1--6.

\bibitem{li2016ternary}
F.~Li, B.~Zhang, and B.~Liu, ``Ternary weight networks,'' \emph{arXiv preprint
  arXiv:1605.04711}, 2016.

\bibitem{meng2017two}
W.~Meng, Z.~Gu, M.~Zhang, and Z.~Wu, ``Two-bit networks for deep learning on
  resource-constrained embedded devices,'' \emph{arXiv preprint
  arXiv:1701.00485}, 2017.

\bibitem{courbariaux2015binaryconnect}
M.~Courbariaux, Y.~Bengio, and J.-P. David, ``Binaryconnect: Training deep
  neural networks with binary weights during propagations,'' in \emph{Advances
  in Neural Information Processing Systems}, 2015, pp. 3123--3131.

\bibitem{rastegari2016xnor}
M.~Rastegari, V.~Ordonez, J.~Redmon, and A.~Farhadi, ``Xnor-net: Imagenet
  classification using binary convolutional neural networks,'' in
  \emph{European Conference on Computer Vision}.\hskip 1em plus 0.5em minus
  0.4em\relax Springer, 2016, pp. 525--542.

\bibitem{courbariaux2016binarized}
M.~Courbariaux, I.~Hubara, D.~Soudry, R.~El-Yaniv, and Y.~Bengio, ``Binarized
  neural networks: Training deep neural networks with weights and activations
  constrained to+ 1 or-1,'' \emph{arXiv preprint arXiv:1602.02830}, 2016.

\bibitem{ImageNet}
O.~Russakovsky, J.~Deng, H.~Su, J.~Krause, S.~Satheesh, S.~Ma, Z.~Huang,
  A.~Karpathy, A.~Khosla, M.~Bernstein, A.~C. Berg, and L.~Fei-Fei, ``{ImageNet
  Large Scale Visual Recognition Challenge},'' \emph{International Journal of
  Computer Vision (IJCV)}, vol. 115, no.~3, pp. 211--252, 2015.

\bibitem{he2016deep}
K.~He, X.~Zhang, S.~Ren, and J.~Sun, ``Deep residual learning for image
  recognition,'' in \emph{Proceedings of the IEEE conference on computer vision
  and pattern recognition}, 2016, pp. 770--778.

\bibitem{lecun1998gradient}
Y.~LeCun, L.~Bottou, Y.~Bengio, and P.~Haffner, ``Gradient-based learning
  applied to document recognition,'' \emph{Proceedings of the IEEE}, vol.~86,
  no.~11, pp. 2278--2324, 1998.

\end{thebibliography}
\bibliographystyle{IEEEtran}

\end{document}